\documentclass[graybox]{svmult}
\usepackage{drum}           
\usepackage{mathptmx}       
\usepackage{helvet}         
\usepackage{courier}        
\usepackage{type1cm}        
\usepackage{amsmath}
\usepackage{makeidx}         
\usepackage{graphicx}        
\usepackage{multicol}        
\usepackage[caption=false,font=footnotesize]{subfig}
\usepackage[bottom]{footmisc}
\usepackage[rightcaption]{sidecap}
\usepackage[square,numbers]{natbib}
\usepackage{algorithm}
\usepackage{algpseudocode}

\makeindex

\begin{document}

\title*{Fundamental Limitations in Performance and Interpretability of Common Planar Rigid-Body Contact Models}
\author{Nima Fazeli, Samuel Zapolsky, Evan Drumwright, and Alberto Rodriguez}

\institute{Nima Fazeli, Alberto Rodriguez \at Mechanical
  Engineering Department, MIT, USA
  \email{$<$nfazeli,albertor$>$@mit.edu} \and Samuel Zapolsky, Evan Drumwright \at
  Toyota Research Institute, CA, USA
  \email{$<$sam.zapolsky,evan.drumwright$>$@tri.global}}

\titlerunning{Limitations in Performance and Interpretability of Planar Rigid-Body Contact Models}

\maketitle

\vspace{-1.0cm}
\abstract{
    The ability to reason about and predict the outcome of contacts is paramount to the successful execution of many robot tasks.
    Analytical rigid-body contact models are used extensively in planning and control due to their computational efficiency and simplicity, yet despite their prevalence, little if any empirical comparison of these models has been made and it is unclear how well they approximate contact outcomes. In this paper, we first formulate a system identification approach for six commonly used contact models in the literature, and use the proposed method to find parameters for an experimental data-set of impacts. Next, we compare the models empirically, and establish a task specific upper bound on the performance of the models and the rigid-body contact model paradigm. We highlight the limitations of these models, salient failure modes, and the care that should be taken in parameter selection, which are ultimately difficult to give a physical interpretation. 
}

\section{Introduction}
Real-time planning, control, and state estimation play a central role in modern robotics. Many contemporary algorithms rely on accurate models of the dynamics of a robot as well as models of interactions of the robot with its environment (\cite{ChavanDafle2015a,Hogan2016,Posa2013}). Dynamic frictional interaction between bodies is important to robotic systems undergoing contact, yet has proven to be a challenging phenomenon to model. Challenges include, but are not limited to, non-smooth dynamics (i.e., transitions from absence of contact to sticking/sliding contact), near-impulsive forces with large magnitudes, local deformations (e.g., difficult to predict elastic/plastic indentations), wave propagation through the bodies (e.g., vibrations along rods from impact), sensitivity to initial conditions, and acting friction---a complex phenomenon that emerges from the interactions between microscopic surface asperities---through necessarily coarse models.

\begin{figure}
    \centering
\includegraphics[width=.9\textwidth]{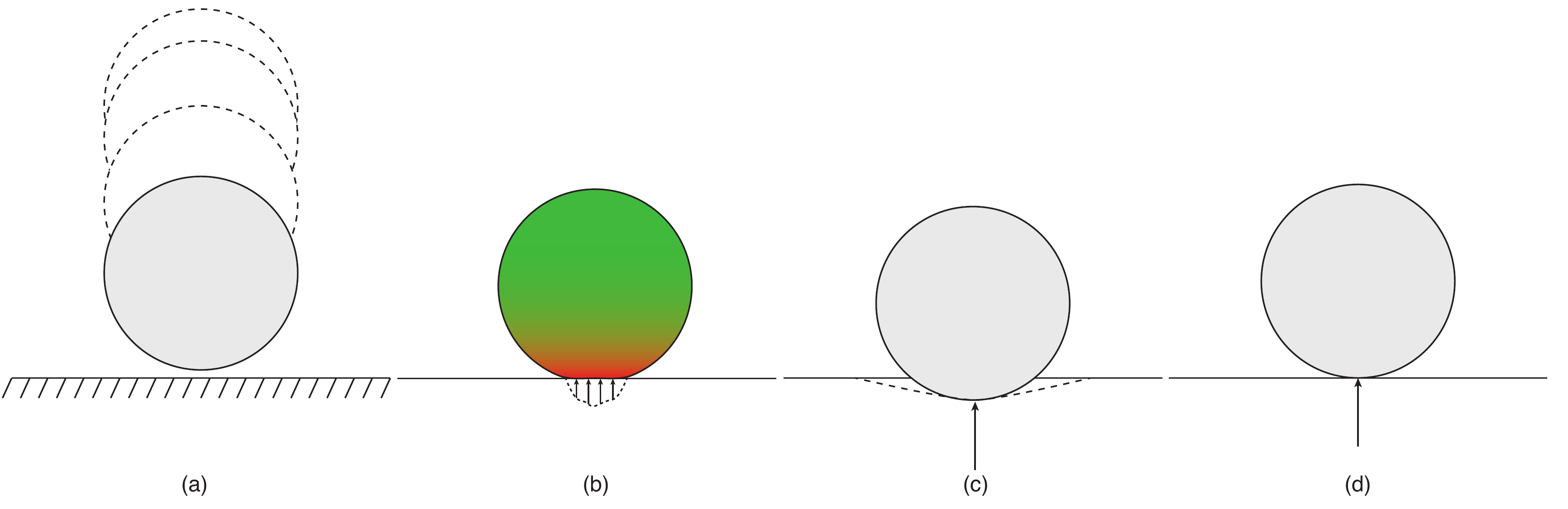}
\caption{Contact paradigms: a) Disk motion prior to contact, b) Elasto-dynamic (colors depict degree of stress), c) Compliant (pseudo-rigid), d) Rigid}
\label{fig:cm_paradigms}       
\end{figure}

Despite these difficulties, various contact models have been proposed and generally follow one of three paradigms: i) fully deformable, elasto-dynamic models; ii) pseudo-rigid, compliant models; and iii) rigid contact. Fig.~\ref{fig:cm_paradigms} shows, conceptually, how each paradigm models contact. 
\vspace{-0.5cm}

\vspace{-0.35cm}
The elasto-dynamic paradigm models deformations for contacting bodies over a finite contact time (i.e., impacts are correctly characterized as non-instantaneous phenomena), can approximately predict elastic/plastic deformations, and are able to model vibrations along the object. These modeling capabilities come at a high computational cost and require identifying greater numbers of system parameters; their greater predictive power is compromised by uncertainty in initial conditions of robotics applications. As such, these models have not been prevalent in robotics, where speed and simplicity play an important role and system identification cannot yet be conducted accurately in real time.

The class of pseudo-rigid, compliant contact models (often designated ``penalty methods'') is the oldest, beginning with the seminal work of Hertz~\cite{hertz1881contact} followed by numerous authors \cite{Kraus97,lankarani1990contact,Marhefka99,Song:2000}. These models assume a rigid core and a thin, deformable layer and apply contact forces that are proportional to the magnitude of deformation. Simulation software such as DANCE \cite{NgThowHing1999DANCEDA} (open-source) and ADAMS \cite{li2001virtual} (proprietary) have been developed under this paradigm. These models are numerically challenging to use in simulation and control applications due to discontinuity in the Coulomb friction model~\cite{Stewart:2000} and nearly rigid contacts. Modeling pseudo-rigid contact from the opposite direction---softening otherwise rigid contacts---was pioneered by Lacoursi\'{e}re~\cite{Lacoursiere:2007} in MathEngine and has been used in numerous rigid body dynamics simulations libraries since (e.g., ODE, Bullet, MuJoCo, DART).

Like full elastodynamic models, pseudo-rigid body models require identifying system parameters (Young's Modulus and Poisson's Ratio); when bodies are known to be nearly rigid (the case of the objects in the present study), the extra flexibility afforded by the pseudo-rigid body models can thus be undesirable.

Rigid contact models resolve contact without permitting geometric intersections between bodies. These methods assume that deformations are nonexistent and contacts ``appear'' discontinuously as a function of configuration---meaning that impacts must be resolved instantaneously. The rigid contact model distinguishes between non-impacting and impacting contacts; while the combination of complementarity conditions, Coulomb friction, and stiction constraints often yields a deterministic model for non-impacting contact, while impacting contact is underdetermined. Rigid impact models effectively add constraints (like restitution) to select points in the feasible impact space (concept described by~\citet{Chatterjee:1998}). Rigid-body impact models span centuries of research, starting with Newton, to contributions from Poisson, and more ``modern" treatments~\cite{Whittaker:1944,Stronge:1990,Wang:1992}. These models typically require 2--3 parameters (often referred to as normal restitution and friction coefficient) and are commonly used in robotic manipulation~\cite{ChavanDafle2015a,Hogan2016} and locomotion~\cite{Posa2013}.

Despite the variety and wide usage of existing rigid impact models, relatively little empirical evaluation and comparison among models has been conducted. The performance of such models has not been quantified; consequently, the limitations of these models are not well understood. Additionally, systematic parameter estimation methodologies have not been developed for the class of rigid impact models. In this paper we empirically evaluate the performance of six common rigid impact models used in robotics on a planar impact experiment, demonstrate that a fundamental upper-limit for performance of these models exists, and provide a systematic methodology for optimal parameter identification to maximize the models' predictive performance. We further show that regions for good/poor predictions can be identified and demonstrate the wide variability in model parameters depending on the choice of data used to identify the models.

\vspace{-0.5cm}
\section{Rigid contact models}
Rigid contact models couple the Newton-Euler equations of motion to contact equations (non-interpenetration constraints, Coulomb friction, stiction, \emph{etc.}) to predict forces imparted to the bodies, and, subsequent motion by extension. The class of rigid contact models makes three simplifying assumptions: i) bodies undergoing contact sustain negligible deformation (coupling this assumption with convex geometry implies that a single point of contact need be considered);
ii) all impacts and their resolutions are instantaneous, and therefore velocity changes discontinuously; and iii) the configuration of impacting bodies does not change during impact (follows from the instantaneity assumption).

\begin{SCfigure}[0.35][h]
\centering
\caption{Planar object making contact with a horizontal surface. The contact point is denoted by $C$.} 
\includegraphics[width=0.65\textwidth]{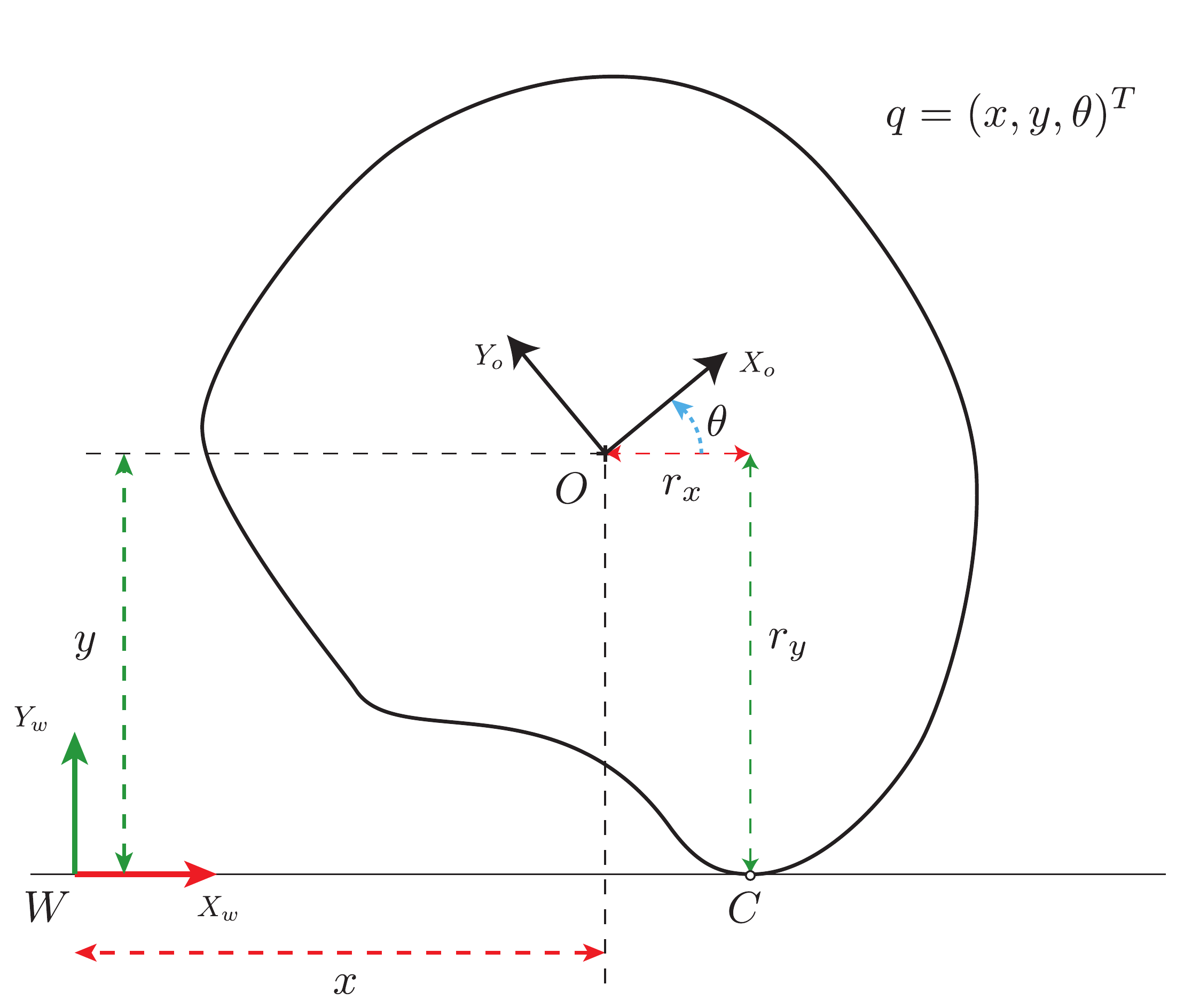} \label{fig:2.1}
\end{SCfigure}

Consider Fig.~\ref{fig:2.1}, where a planar object makes contact with a fixed horizontal surface. The configuration of the object is $\vect{q}_o \equiv \tr{\begin{bmatrix}x & y & \theta\end{bmatrix}}$, the velocity $\vect{v}_o \equiv \dot{\vect{q}}_o$, the point of contact $\vect{r} \equiv \tr{\begin{bmatrix} r_x & r_y \end{bmatrix}}$, and the mass and second moment of inertia $m$ and $I$ respectively. Unless stated otherwise, all points and vectors defined in a ``world'' (stationary) reference frame. We can represent a rigid model for contact of a planar moving body with a fixed surface as a mapping:
\begin{align}
    f_{c}:\{\vect{q}_o,\vect{v}_o^i, \vect{r},m,I\} \rightarrow \{\vect{v}_o^f\}
\end{align}
where the superscripts $i$ and $f$ denote the pre- and post-contact values (recall that the rigid body assumption, which implies instantaneity, entails that $\vect{q}_o^i = \vect{q}_o^f$, so we omit such subscripts from $\vect{q}_o$). Under the rigid-body assumption, Newtonian mechanics  relates the pre- and post-contact states by:
\begin{align}
    \mat{M} \frac{\vect{v}_o^f-\vect{v}_o^i}{dt} = \tr{\mat{J}(\vect{q}, \vect{r})} \vect{F}_c \;\; \rightarrow \;\; \vect{v}_o^f = \vect{v}_o^i + \inv{\mat{M}}\tr{\mat{J}}\vect{P}
\end{align}
where $\mat{M} \in \mathbb{R}^{3 \times 3}$ is the generalized inertia matrix of the rigid body,  $\vect{P}=\vect{F}_c\ dt$ denotes the impulse imparted to the bodies due to contact ($\vect{F}_c$ is the non-impulsive version of this force), and $\mat{J} \in \mathbb{R}^{2 \times 3}$ denotes the Jacobian matrix that maps generalized velocities to contact-frame velocities and contact-frame impulses to generalized impulses (via the transpose of the Jacobian). Note that this relation holds for the moment before and after contact, but is insufficient to determine the contact impulse (the vector operations above yield three equations and six unknowns: $\vect{v}_o^f$ and $\vect{P}$).

Given the unknown imparted impulse we can solve for the post-contact velocity; choosing the imparted impulse is precisely the job of the contact models. To gain an understanding of how contact models choose $\vect{P}$, in sec.~\ref{sec:fundConst} we will briefly discuss the fundamental constraints of contact. This understanding will lay the foundation to the system identification formulation as well as the data-driven methods.
\vspace{-0.65cm}

\vspace{-1.0cm}
\subsection{Fundamental constraints of contact} \label{sec:fundConst}
Under the rigid-body assumption, all contact must obey three fundamental constraints: i) only compressive impulses may be applied along the surface normal (contacts are not permitted to pull); ii) impacting bodies must not penetrate; and iii) the total system momentum must be conserved. To obtain a visual representation of the feasible impulse space for 2D contact we adopt the treatment discussed in~\cite{Chatterjee:1998}. We write the change in momentum of the object due to contact as:
\begin{align}
    \vect{v}_o^f-\vect{v}_o^i = \inv{\mat{M}}\tr{\mat{J}}\vect{P}
\end{align}
We write the change in velocity of the contact point by pre-multiplying both sides of the expression by $\mat{J}$:
\begin{align} \label{eq:impact_eq}
    \mat{J}(\vect{v}_o^f-\vect{v}_o^i) = \mat{J}\inv{\mat{M}}\tr{\mat{J}} \vect{P} \quad  \rightarrow \quad (\vect{v}_c^f-\vect{v}_c^i) = \mat{M}_c^{-1} \vect{P}
\end{align}
Here $\vect{v}_c$ is the $2\times 1$ velocity vector of the point of contact $c$, $\inv{\mat{M}_c} \equiv \mat{J}\inv{\mat{M}}\tr{\mat{J}}$ is the $2\times 2$ \emph{contact space compliance matrix} (making $\mat{M}_c$ the \emph{contact space inertia matrix}. We note that the kinetic energy of the system post-contact may not exceed that of the energy prior to contact; we may write this constraint as:
\begin{align}
    \tr{\vect{v}_c^f}\mat{M}_c\vect{v}_c^f = \alpha \tr{\vect{v}_c^i}\mat{M}_c\vect{v}_c^i, \quad 0 \leq \alpha \leq 1
\end{align}
Recall that the position of the object does not change during an instantaneous impact, so potential energy remains the same. We re-write this expression using~\eqref{eq:impact_eq}, and after expansion and substitution:
\begin{align}
\tr{(\vect{P}+\mat{M}_c\vect{v}_c^i)}\inv{\mat{M}}_c(\vect{P}+\mat{M}_c\vect{v}_c^i) = \alpha  \tr{\vect{v}_c^i}\inv{\mat{M}}_c\vect{v}_c^i
\end{align}
The representation of the expression above is that of an ellipse in 2D due to the positive definite nature of $\mat{M}_c$; the term $\mat{M}_c\vect{v}_c^i$ simply acts to translate the ellipse $\tr{\vect{P}}\inv{\mat{M}}_c\vect{P}$ in $\vect{P}$. The union of all ellipses for all $0 \leq \alpha \leq 1$ gives what we call the Energy Ellipse~\cite{Chatterjee:1998}. 

\begin{SCfigure}[.675][h]
\centering
\caption{The Energy Ellipse: This ellipse is constructed from application of the conservation of energy law before and after the contact event. Points within the ellipse satisfy this law, and points in the shaded region satisfy the non-negative normal velocity component of the contact point.} 
\includegraphics[width=0.6\textwidth]{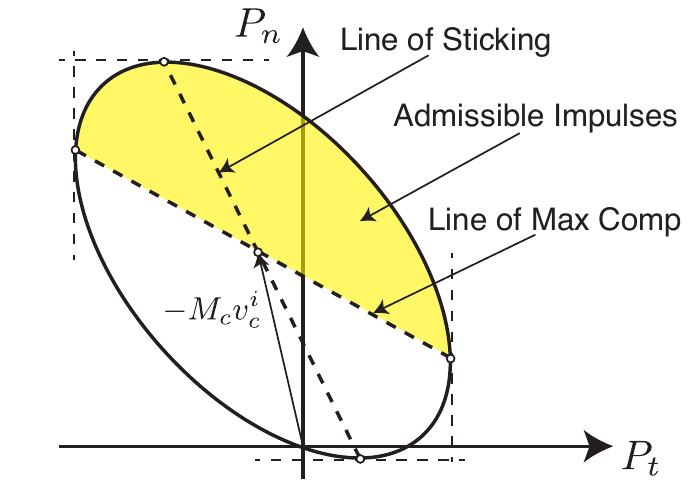} \label{fig:sec_2_energy_ellipse}
\end{SCfigure}

Fig.~\ref{fig:sec_2_energy_ellipse} depicts an Energy Ellipse, where the shaded region is the admissible region based on the fundamental constraints of rigid contact. Two important features of the Energy Ellipse are the lines of sticking and maximum compression. 

\vspace{-0.5cm}

\vspace{-0.5cm}
The line of maximum compression delimits the region for which the separation velocity of the contact point changes sign; regions on and above this line obey the fundamental contact constraints. Points along the line of sticking signify contacts for which the relative tangential velocity of the contact point is zeroed due to friction. The intersection of these two lines is the impulse imparted that eliminates all relative velocity between the bodies at the contact point. The diagram of the Energy Ellipse graphically shows that the three fundamental constraints governing the impact are not sufficient to select an impulse to resolve the impact. Contact models introduce additional assumptions and constraints to choose a unique point within the feasible space.  

In the next section we briefly introduce the models considered in this paper, and demonstrate how they select points within the Energy Ellipse.

\vspace{-0.5cm}
\subsection{Rigid contact models studied in this paper}
The purpose of a rigid contact model is to select a point in the feasible impulse space demarcated with the Energy Ellipse. In this section we very briefly introduce the six models we used in this study; and refer the interested reader to~\cite{Fazeli2017contact} for further details. The six models studied are all parameterized by two variables, denoted as $(\mu,\epsilon)$ and often referred to as the coefficient of friction and restitution, respectively. When the contact-space compliance matrix is diagonal, the former is used to regulate the magnitude of tangential momentum of the contact point and the latter is used to regulate the normal component. 

The first two models are \textbf{Anitescu-Potra Newton}~\cite{Anitescu:1997} and \textbf{Anitescu-Potra Poisson}~\cite{Anitescu:1997}, which are closely related to Stewart-Trinkle~\cite{Stewart:1996}, differing primarily in the restitution used for contact resolution as their names suggest. The \textbf{Drumwright-Shell}~\cite{Drumwright:2010b} also uses a Poisson-type contact model that computes contact forces that maximize kinetic energy dissipation. \textbf{Mirtich}~\cite{Mirtich:1996vt} is an incremental collision model and a computational implementation of Stronge's work~\cite{Stronge:1990} relating physical work done in the compression and restitution phases of an impact. \textbf{Wang-Mason}~\cite{Wang:1992} relates the magnitude of the impulses in the compression and restitution phases of an impact using a Poisson-type restitution. \textbf{Whittaker}~\cite{Whittaker:1944,Kane:1985} is an algebraic model that requires solution to a nonlinear system of equations and both 2D and 3D versions have been studied. 

We now define two post hoc models, the \textbf{Best Post Hoc} and the \textbf{Ideal Rigid-Body Bound (IRB Bound)}. The Best Post Hoc model selects the best predicting rigid-body contact model, of the six considered in this paper, after each impact event, and the IRB Bound selects the best predicting impulse from the Energy Ellipse. We emphasize that the IRB Bound is able to select any possible feasible impulse, but the Best Post Hoc only selects a model (spanning a sub-space of the Energy Ellipse). Consequently the IRB Bound quantifies the absolute best performance feasible for any possible rigid-body contact model, and the Best Post Hoc quantifies the best possible performance for the models studied. Neither model is suitable for making predictions, since each makes their choice post hoc.



\begin{figure}
    \centering
    \includegraphics[width=0.9\textwidth]{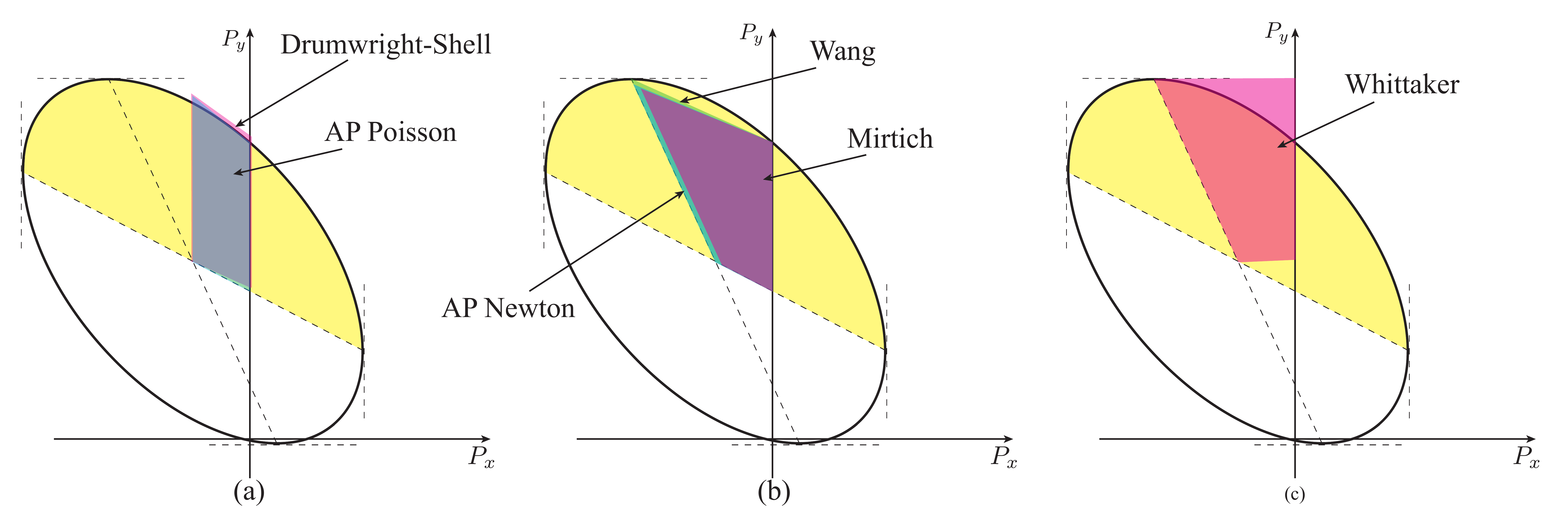} 
    \caption{Regions of the admissible contact space covered by each model. Note the similarity between the models and the limited coverage of the admissible space.} \label{fig:energy_ellipse_regions}
\end{figure}

Fig. \ref{fig:energy_ellipse_regions} depicts the admissible contact regions for each model, grouped based on graphical similarity. This similarity is not necessarily indicative of the way the regions are constructed. These regions are the union of all predicted impulses for all admissible values of the model parameters.

Fig. \ref{fig:energy_ellipse_regions}~(a) shows that the Drumwright-Shell and AP Poisson models cover very similar regions of the space. Fig.~\ref{fig:energy_ellipse_regions} also provides intuition for the definitions used in contact parameters. For example, the coefficient of restitution regulates the magnitude of the normal impulse; therefore, models with vertical side boundaries use the ratio of the post to pre-contact normal momenta. Fig. \ref{fig:energy_ellipse_regions}~(b) depicts the regions from the AP Newton, Wang-Mason, and the Mirtich models. We note that the definition of the coefficient of restitution used in these three models relates the relative normal velocities pre- and post-contact, resulting in fairly similar regions. Fig. \ref{fig:energy_ellipse_regions}~(c) depicts the Whittaker model, for which the horizontal boundaries of the feasible region imply that the coefficient of friction relates the tangential momenta pre- and post-contact, differing significantly in interpretation from the previous models. 

Two key observations are: i) no single model is able to span the full admissible region of contact (we further note that the regions covered by the various models are quite similar in shape); ii)  any given model will make predictions only up to the line of sticking, but not beyond it; therefore the models are not able to predict ``back-spin" (i.e., the tangential velocity of the contact point post-contact changes direction). This issue is sometimes alleviated by contact models that utilize an additional tangential restitution, but the physical interpretation and utility of such restitution is debatable \cite{Chatterjee:1998}.



\vspace{-0.5cm}
\section{System identification formulation and model evaluation}
\label{sec:problem_formulation}
System identification applied to inertial parameter estimation of robotic systems has received considerable attention within the robotics research community~\cite{fazeli2016parameter,Fazeli2016,Gautier1988,Khosla1985,Koval2015} and several practical implementations have been developed and applied to industrial robots~\cite{nubiola2013absolute}. Control architectures have also been developed to estimate and adapt to uncertainties over these dynamic parameters ~\cite{Hogan2016,Slotine1987}, and have been shown to exhibit asymptotic convergence to desired trajectories in the case of free space motion of manipulators.

System identification is a necessary step to maximize the predictive performance of these models, yet to our knowledge no systematic formulation of this step has been proposed. To this end, we first propose a system identification formulation in impulse space, leveraging the Energy Ellipse. Next, we briefly discuss the experimental setup and data collection protocols used in this study; we then apply the system identification procedure to find the optimal parameters for each of the studied models, allowing us to empirically compare predictions and use the IRB Bound and Best Post Hoc models to establish upper limits in predictive performance. Finally, we study individual experimental trials and demonstrate the large variability in model parameters across experimental trials (indicating challenges in interpreting parameters). We also quantify the predictive range of each model by evaluating the number of outcomes across individual instances of the experiment that can be explained by \emph{any} choice of parameter set. 

To perform system identification, we first compute the impulse imparted on the object from pre- and post-contact states (\citet{fazeli2016parameter}) under the ideal rigid body impact model by solving:
\begin{align}
    \vect{P}_{IRB} = & \; \underset{\vect{P}}{\text{arg min}}  \; \; || \mat{M}(\vect{v}_o^f - \vect{v}_o^i) - \mat{J}^T \vect{P}||_2
\end{align}
Then we can identify the parameters for a single impact by solving the following optimization program:
\begin{align}
    (\mu^*,\epsilon^*) = & \; \underset{(\mu, \epsilon)}{\text{arg min}}  \; \; ||\vect{P}_{IRB} - \text{Model}(\mu, \epsilon)||_2 \\
    & \text{s.t.} \quad  \;  0 \leq \mu \leq \mu_s \quad \; \; 0 \leq \epsilon \leq 1
\end{align}
The optimization program is convex within the constraints (as will be explained in more detail in Sec.~\ref{sec:best_avg}) and takes place in the space of imparted impulses and the Energy Ellipse. The upper bound on the coefficient of friction $\mu_s$ is the lowest value of the coefficient of friction for which the model predicts sticking contact and can be found with a line search using a bisection method. For values of $\mu > \mu_s$ and a fixed $\epsilon$, the predicted impulse from the model will remain the same and the cost function will not have a gradient, making the optimization ill-posed numerically. This implies that pre and post contact conditions exist which do not provide sufficient information to estimate $\mu$ correctly and render this parameter unobservable, more on this in Sec.~\ref{sec:best_ind}. 

To perform system identification for a batch of impacts, we simply sum up the cost function over the set of impacts (Sec.~\ref{sec:best_avg}).

\vspace{-0.75cm}
\subsection{Experimental data collection} \label{sec:data}
\begin{SCfigure}[0.8][h]
    \includegraphics[width=0.650\textwidth]{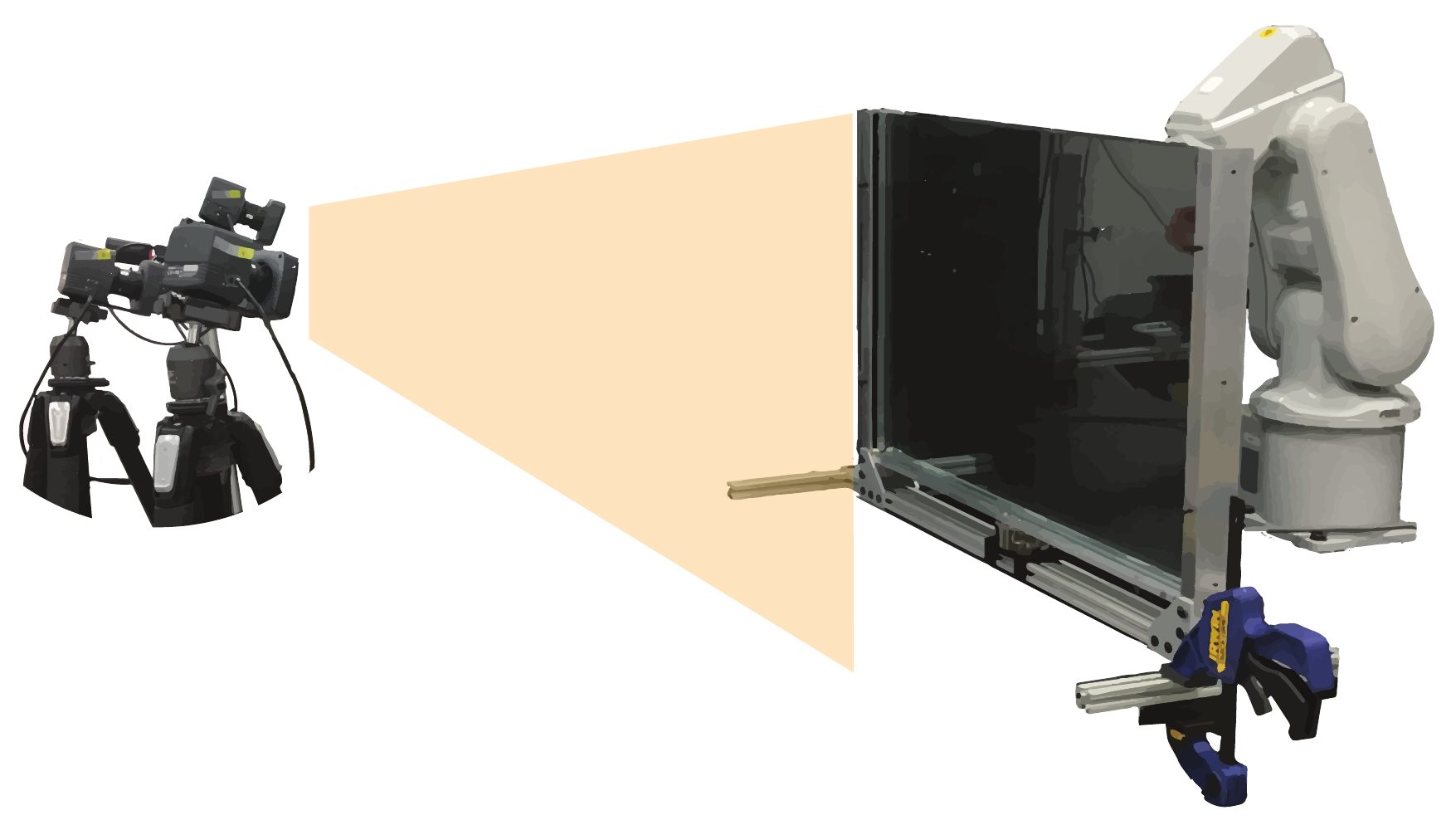}
    \caption{Experimental setup: Autonomous data collection for planar impact data. A planar object is constrained to move between two lubricated vertical glass planes. The robot uses a directional magnetic latch to pick up the object across the glass and provide control over position and velocity of the object. The Vicon motion capture system is used to track the states of the object.}
    \label{fig:exp_setup}
\end{SCfigure}

In order to evaluate and compare the contact models we conceived the experimental setup depicted in Fig.~\ref{fig:exp_setup} \cite{Fazeli2017contact,fazeli2016parameter}. The purpose of the apparatus is to autonomously collect measurements of the trajectory of falling planar objects initialized from a specifiable set of initial conditions. We chose this experiment because it captures key features of a dynamic frictional interaction while being amenable to careful monitoring and instrumentation resulting in high quality measurements; for further details see~\cite{Fazeli2017contact}.

The experimental setup is composed of a ``dropping arena" designed to constrain the motion of planar objects to a plane using lubricated glass planes. The planar objects are equipped with markers to be tracked using the Vicon motion capture system at 250 Hz, and directional magnets allowing the robot to latch onto the object and prescribe desired initial state to the object for each drop.

\begin{figure}
    \centering
    \includegraphics[width=0.6\textwidth]{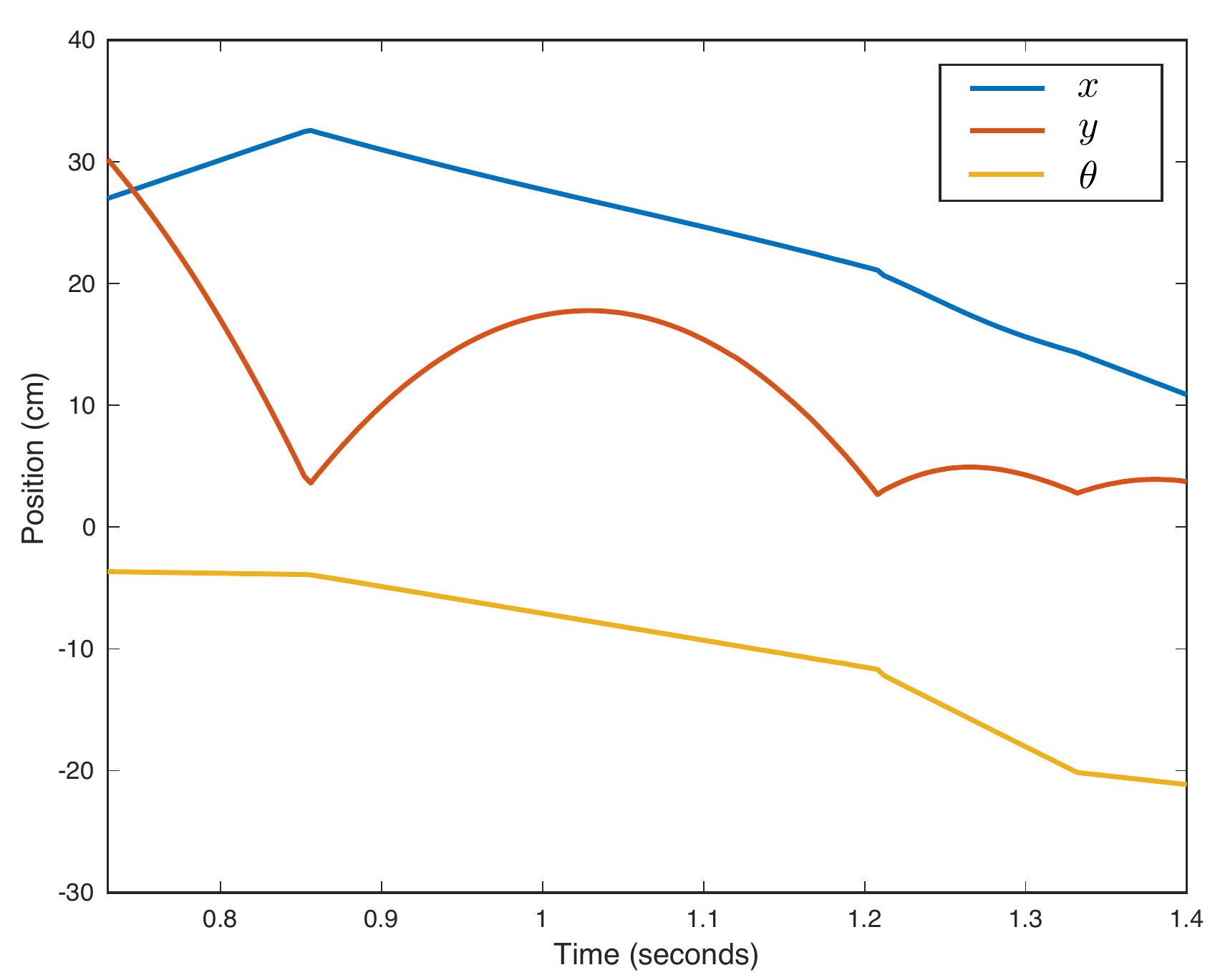}
    \caption{The configurations of the object as a function of time during an experimental run. $\theta$ is in radians but has been scaled by the radius of gyration of the object.}
    \label{fig:Trace}
\end{figure}

We collected a total of 2,000 drops. In some instances the tracking system may lose frames or suffer spurious errors due to reflections; to detect and remove these effects each experimental run was subjected to a series of tests to validate the correctness of the measurements. The three most important tests were: i) the energy test: verifies constant energy during ballistic motion and energy loss during times of contact, ii) a frame drop test, and iii) a deviation test: checks deviations of trajectories during ballistic motion from parabolic paths. A total of 1,718 of the 2,000 drops passed all tests, and we used this set for the rest of this study. An example of the recorded configurations for one experiment is shown in Fig.~\ref{fig:Trace}.

To detect the time of contact we use the fact that the object undergoes ballistic motion, consequently the acceleration of the object is smooth. Any impact event induces a spikes in the second time derivative of the state vector, and we use a threshold on these spikes to detect events. We further use the ballistic motion of the object to fit parabola to the pre- and post-contact trajectories, thereby mitigating the effects of noise on the state measurements and yielding better approximations to velocities of the object; for more details see \cite{Fazeli2017contact}.

\vspace{-0.5cm}
\subsection{Results of system ID and model prediction performance} 
We conduct the identification in two scenarios, once in batch (Sec.~\ref{sec:best_avg} ensemble parameters), and once for individual impacts (Sec.~\ref{sec:best_ind} individual best parameters). The former scenario will allow us to evaluate the performance of the models in predicting outcomes on previously unseen data, and we will use the latter to show the large variation in parameters across impacts and limitations in the predictions made by the models.

\vspace{-0.75cm}
\subsubsection{Ensemble parameter identification} \label{sec:best_avg}
\begin{figure}[h]
  \centering
  \subfloat[Objective function (Whittaker model).]
  {\includegraphics[height=1.55in]{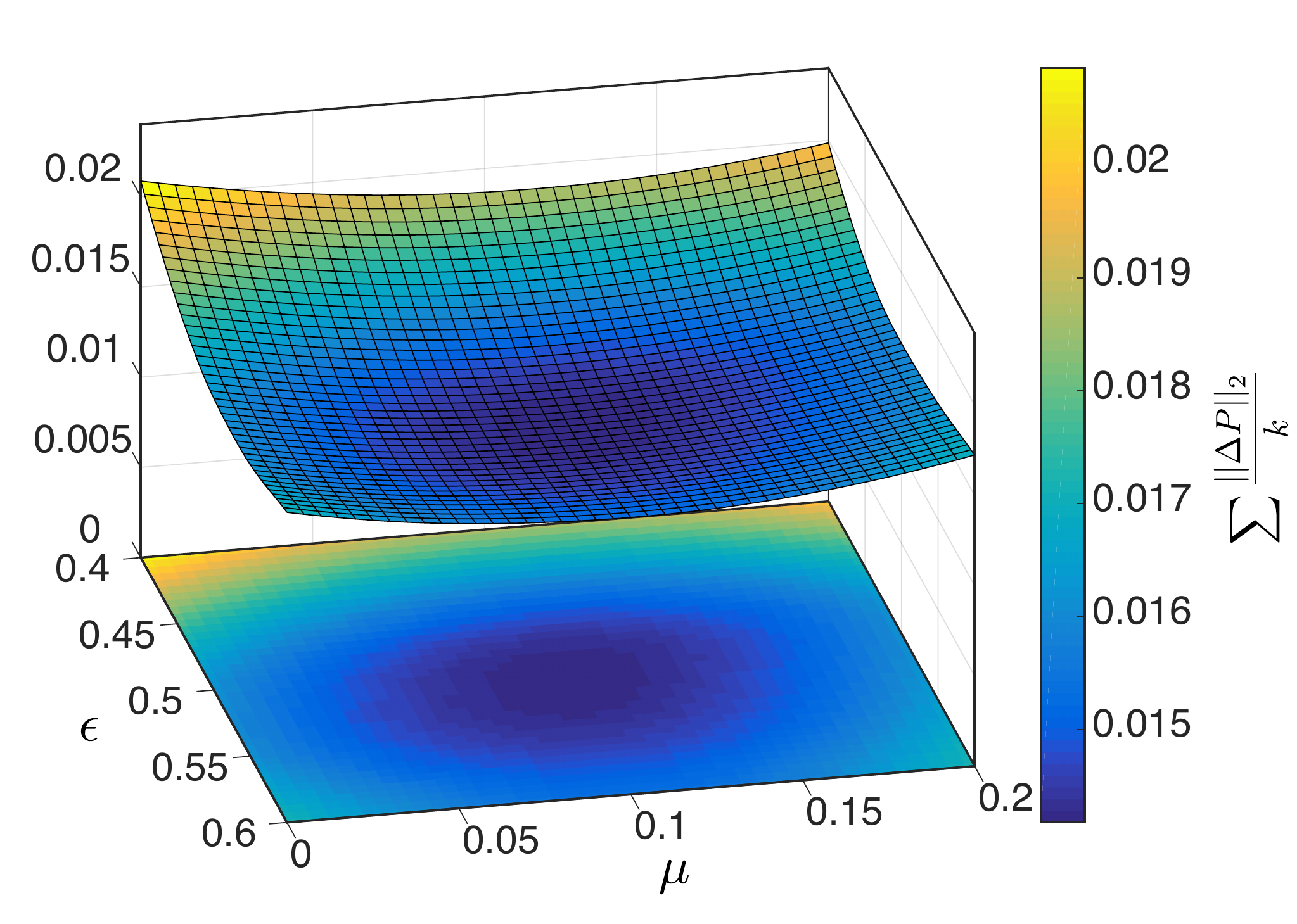}} ~
  \subfloat[Model parameters Convergence.]
  {\includegraphics[height=1.55in]{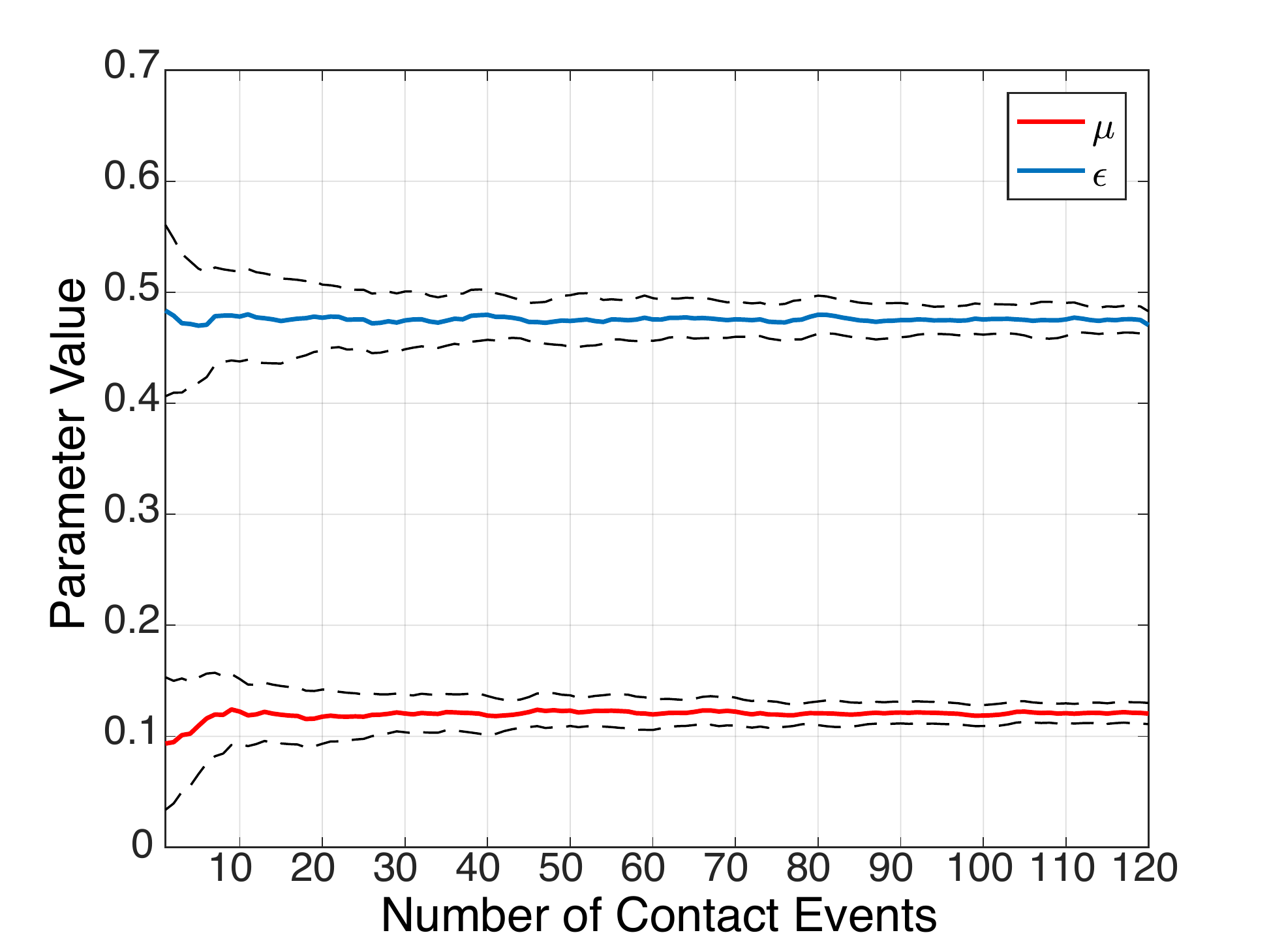}}
  \caption{a) Cost function plotted as a function of the parameters for the Whittaker model for 120 drops. b) Convergence behavior of the model parameters as a function of number of contacts used for the identification, the solid colored lines are the average parameter values over a random set of 20 samples and the dashed black lines depict 1 standard deviation.}
  \label{fig:whittakerEnsamble}
\end{figure}

Ensemble parameters were identified from a randomly selected subset of the data as per the procedure described in Section~\ref{sec:problem_formulation}. Fig.~ \ref{fig:whittakerEnsamble}~(a) shows the shape of the cost function. Thifunction is convex, which is to be expected since the region of admissible impulses predicted by all models is convex and the values of the predicted impulses grow proportionally to the magnitudes of the parameters. Intuitively, if there exists a pair of parameters for a given impact that perfectly predicts the impulse, then the cost associated with that pair would be zero and all other points within the convex set would have error values that grow proportional to deviations from the optimal pair. For a point outside the set of admissible impulses---due to the imposed constraint $\mu \leq \mu_s$---then  there is a point that lies on the boundary of the set for which the distance takes on a minimum.


Fig. \ref{fig:whittakerEnsamble} (b) shows the convergence of the parameters as a function of the number of data used for the Whittaker model; for every $k$ contact events we selected 20 sets of size $k$ at random without replacement from the experiments. Tab.~\ref{tab:average_best} lists the results of the identification for all 6 models for $k=120$. 

\begin{figure}
    \centering
    \includegraphics[width=0.7\textwidth]{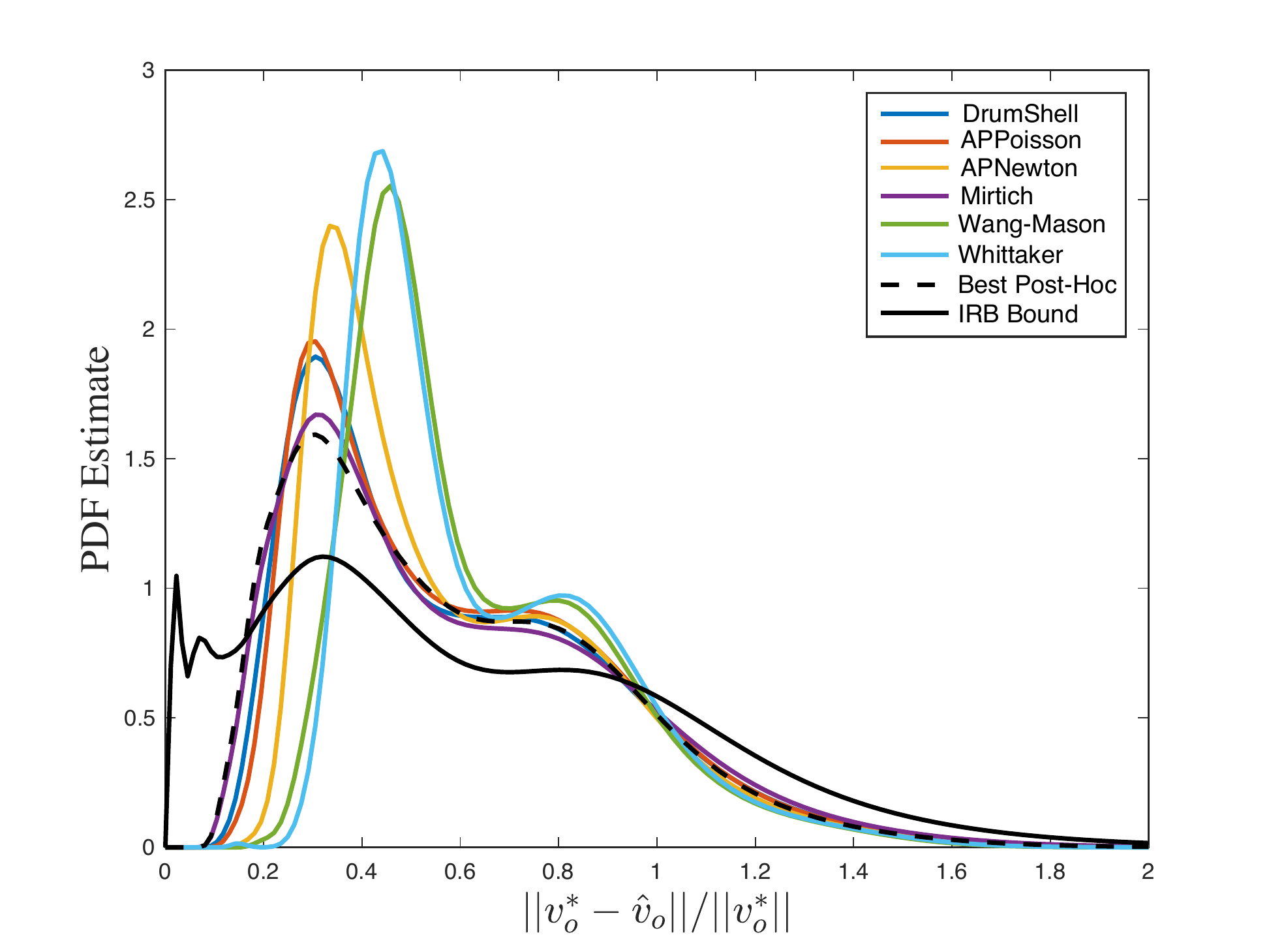}
    \caption{Estimated probability density functions (PDFs) of the $\ell_2$-norm error in post-contact velocity of the center of mass.}
  \label{fig:norm}
\end{figure}

Tab.~\ref{tab:average_best} shows that the values for the coefficient of friction found for each model are much lower than the value (0.3) estimated from an inclined surface experiment between PLA (material of the object used) and Aluminium (material of the impact plate). We believe that the major contributing factor to the discrepancy is the complex interaction that the coefficient of friction tries to model in the highly dynamic dropping experiments. The object can undergo a number of rich and difficult to model interactions in a very short period of time, such as small local deformations, finite-time impact, vibrations, and impulsive torsional frictional forces (in conjunction with the modeled linear ones). The use of just two parameters, which in this case are state independent, is not sufficient to describe the resulting behavior, and the identified parameters lose meaning as they are compensating for all the unmodelled dynamics. In contrast, the inclined surface experiment is slow and controlled, and one parameter is more descriptive of the observations. This is a known phenomenon,~\cite{Chatterjee:1998} discusses the simplicity of impact models, in that they attempt to capture a complicated phenomenon in a minimalistic way and with intuitive parameters. This intuition is valuable: it gives a direct interpretation to the way these parameters affect the outcome of a prediction, but it is unreasonable to expect the values found for these parameters to match those found from other experiments. The coefficient of friction in this context is not necessarily the same as that measured from the inclined surface experiment, though they have the same name; this thought will be elaborated upon in Sec.~\ref{sec:best-param}. 
It is important to note that using the inclined surface experimental value for $\mu$ leads to significantly poorer predictive accuracy. 
\vspace{-0.5cm}
\begin{table}[]
    \centering
    \caption{Average-best parameter identified values and the percentage a model was chosen as the best and worst model}
    \setlength\tabcolsep{0.45cm}
    \begin{tabular}{| c || c | c || c | c |} \hline 
        \textbf{Model}  &   $\mu$    & $\epsilon$ &   \textbf{Best\%} & \textbf{Worst} \%   \\ \hline \hline
        DrumShell   &   $0.081 \pm 0.008$    & $0.516 \pm 0.009$  &   4  & 11      \\ 
        AP Poisson  &   $0.101 \pm 0.007$    & $0.526 \pm 0.008$  &   7  & 6      \\ 
        AP Newton   &   $0.084 \pm 0.007$    & $0.547 \pm 0.009$  &   25 & 0      \\ 
        Mirtich     &   $0.062 \pm 0.009$    & $0.558 \pm 0.010$  &   17 & 43      \\
        Wang-Mason  &   $0.120 \pm 0.008$    & $0.537 \pm 0.010$  &   16 & 10      \\
        Whittaker   &   $0.111 \pm 0.008$    & $0.484 \pm 0.011$  &   31 & 30      \\ \hline
    \end{tabular}
    \label{tab:average_best}
\end{table}
\vspace{-0.2cm}
Fig.~\ref{fig:norm} depicts the estimated PDF of the $\ell_2$-norm error of predicted $\vect{v}_o$ (we normalize the angular velocity by the radius of gyration so that the linear and angular terms are of the same order). The Best Post Hoc model is also plotted to show a performance upper bound on the analytical model; that is to say, if, for each trial, we knew \textit{a priori} which model would perform best. Tab.~\ref{tab:average_best} shows the number of times each model was chosen as the best and worst performing for the dataset. We can see that the Whittaker model is quite erratic, while the most consistent model is the AP Newton model, with a significant number of best predictions and no worst. 

The IRB Bound model is the result of directly searching for the best impulse that explains the difference between pre- and post-contact velocities subject to the constraints of rigid-body point contact. This post hoc model quantifies the absolute best possible performance for any rigid-body contact model under the assumption of point contact.

\begin{SCfigure}[0.3][h]
    \includegraphics[width=0.7\textwidth]{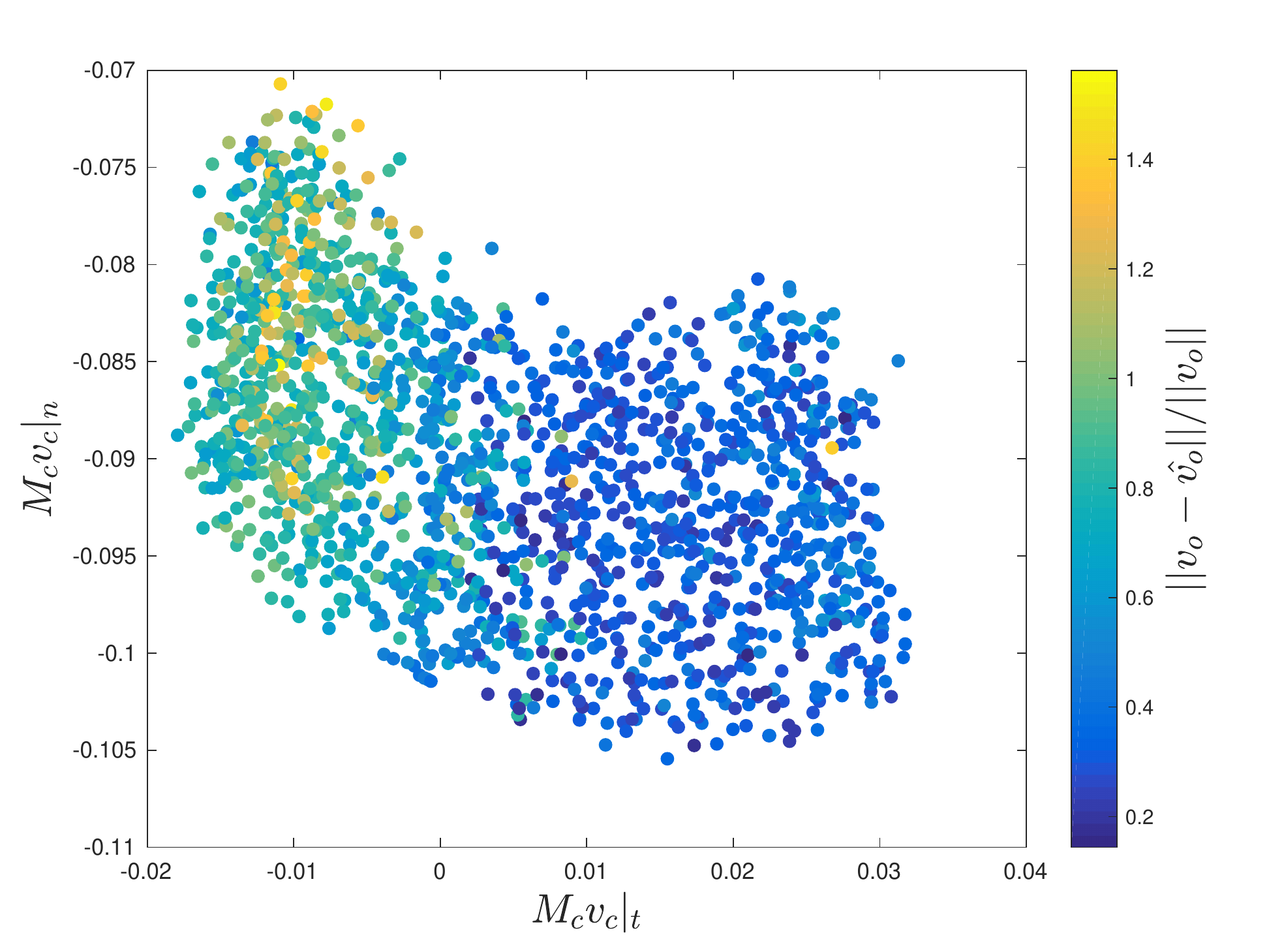}
    \caption{Performance of AP Newton model depicted in the momentum space as a function of the normal and tangential components of the pre-contact momentum of the system measured at the contact point. Blue regions represent high accuracy, yellow regions predict high error.}
    \label{fig:safe_regions}
\end{SCfigure}

It is important to note that the error introduced by the rigid body assumptions is not distinguishable from error resulting from uncertainty in sensor measurements, but our setup was designed to minimize the latter errors. The implication of this ambiguity is that it may be impossible to construct a zero error model without violating the rigid body or rigid  contact models, meaning that the best we can hope to achieve in terms of performance is the ``Ideal'' line.
\vspace{-0.5cm}
\vspace{-0.35cm}
We can further examine the errors made in the models' predictions to quantify regions of the input space in which we expect models to perform comparatively better and worse. This information is valuable if we are interested in selecting initial conditions that are easier to predict or rejecting initial conditions that may be more difficult to predict (for purposes of control). Fig.~ \ref{fig:safe_regions} depicts the performance of the AP Poisson model in predicting the outcome of a contact (error in predicting the velocity of the COM) as a function of the normal and tangential components of the pre-contact system momentum measured at the contact point. The region in which the tangential momentum at the contact point is near zero or changes direction proves to be the most difficult to predict; we will discuss this phenomenon in Section~\ref{sec:best_ind}.

\vspace{-0.5cm}
\subsubsection{Parameter identification for individual impacts} \label{sec:best_ind}

The optimization in Section~\ref{sec:best_avg} to find the ensemble parameters hides some of the details of individual trials. In this section we find the values of $(\mu,\epsilon)$ that best explain each drop, for example Fig.~\ref{fig:best_individual} shows the identified parameters for the AP Newton model, and we make the following observations:
\begin{figure}[h]
  \centering
  \includegraphics[width=0.9\textwidth]{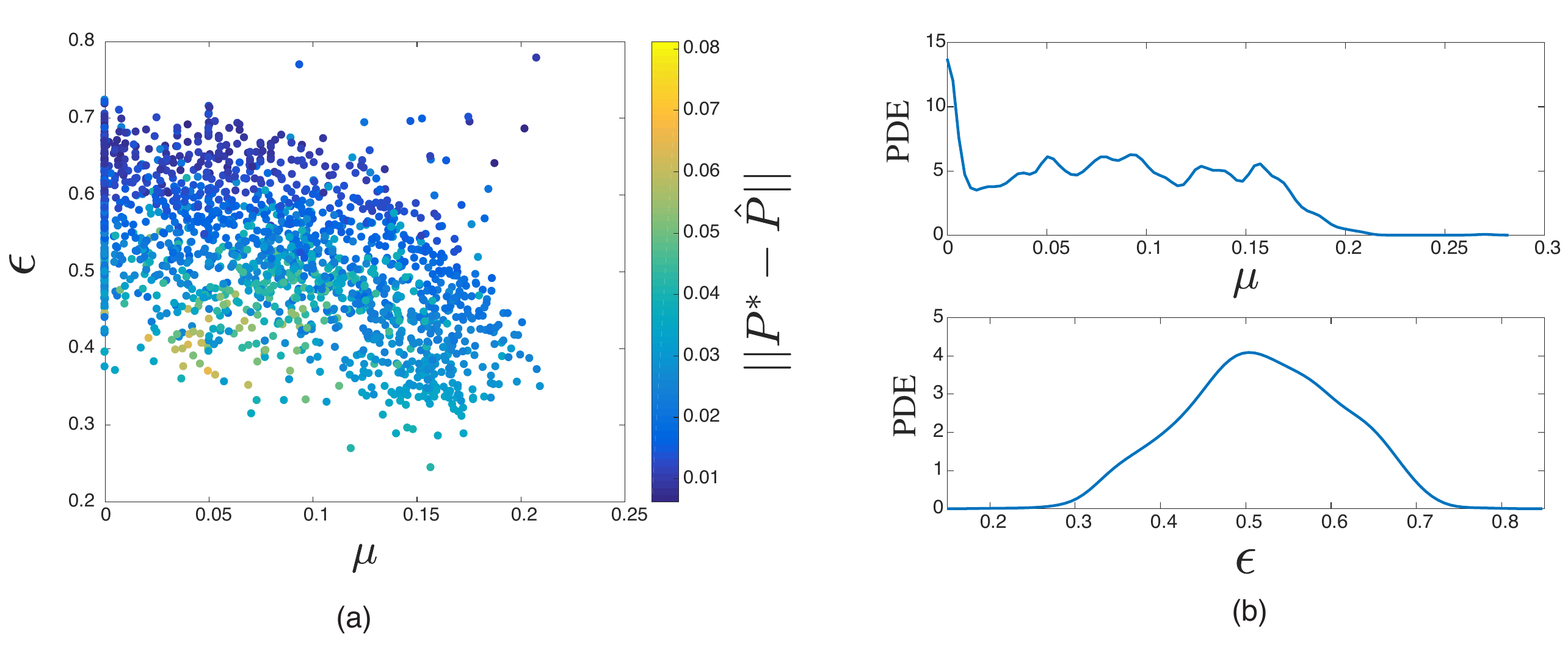}
  \caption{Identified values of $(\mu,\epsilon)$ for each drop, where a) parameter values colored by the ability of the identified pair to predict the impulse, and b) distribution of the parameters.}
  \label{fig:best_individual}
\end{figure}
\vspace{-0.5cm}
\begin{enumerate}
    \item We observe a sharp peak at $\mu=0$.
    \item The lower left region in Fig.~ \ref{fig:best_individual}a shows a loss in performance.
    \item The distribution of $\epsilon$ is uni-modal.
    \item The means of the distributions of $\mu$ and $\epsilon$ are very close to the ``optimal'' values computed for the ensemble parameters.
    \item Neither distribution is tightly localized about a single value. This is perhaps the most important feature to be noted since it is common practice to select a single value of the parameters.
\end{enumerate}

\label{sec:best-param}
The fact that such variation in parameter values exists, particularly for the value of $\mu$, begins to highlight issues of interpretation of the parameters. The intuitive expectation for a coefficient of friction is that it should be constant for all impacts, but these experiments demonstrate that this is not the case. The coefficient of friction in these models acts more similarly to a restitution variable that regulates the magnitude of the tangential impulse as a function of the pre-contact conditions rather than as the typical notion of the coefficient of friction (e.g., that used in the inclined surface experiments).

Part of the challenge in identifying $\mu$ is that not all impacts are informative. Impacts that stick provide no information other than a lower bound for $\mu$. Effectively, any coefficient of friction above the value found by the identification is also able to explain the interaction equally well which partially explains the dispersion in identified $\mu$. To prevent parameter drift and an ill-posed optimization, we use $\mu_s$ as a bound on the coefficient of friction. 

\begin{figure}[h]
  \centering
  \subfloat[Configuration 1]
  {\includegraphics[width=0.45\textwidth]{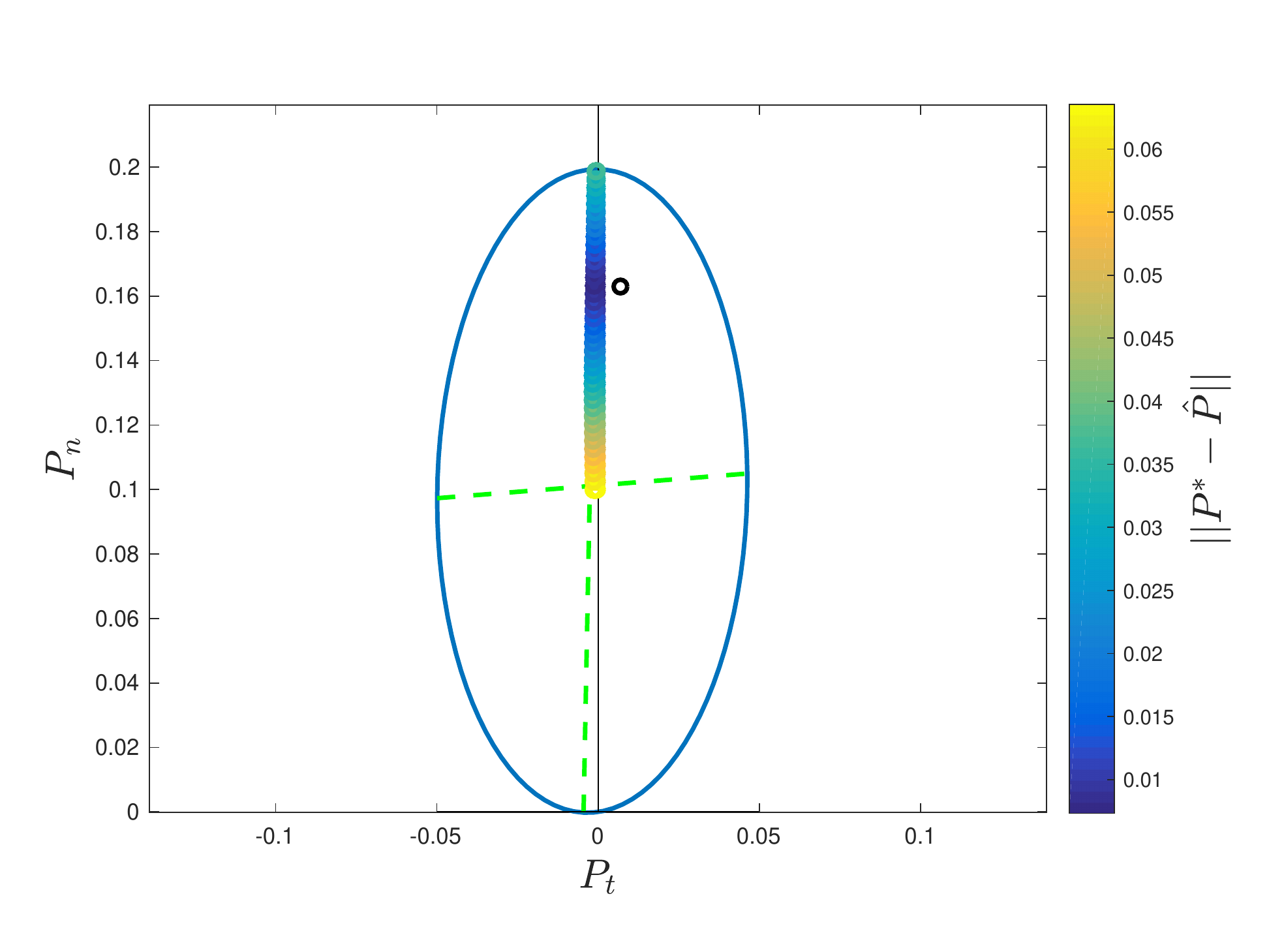}} ~
  \subfloat[Configuration 2]
  {\includegraphics[width=0.45\textwidth]{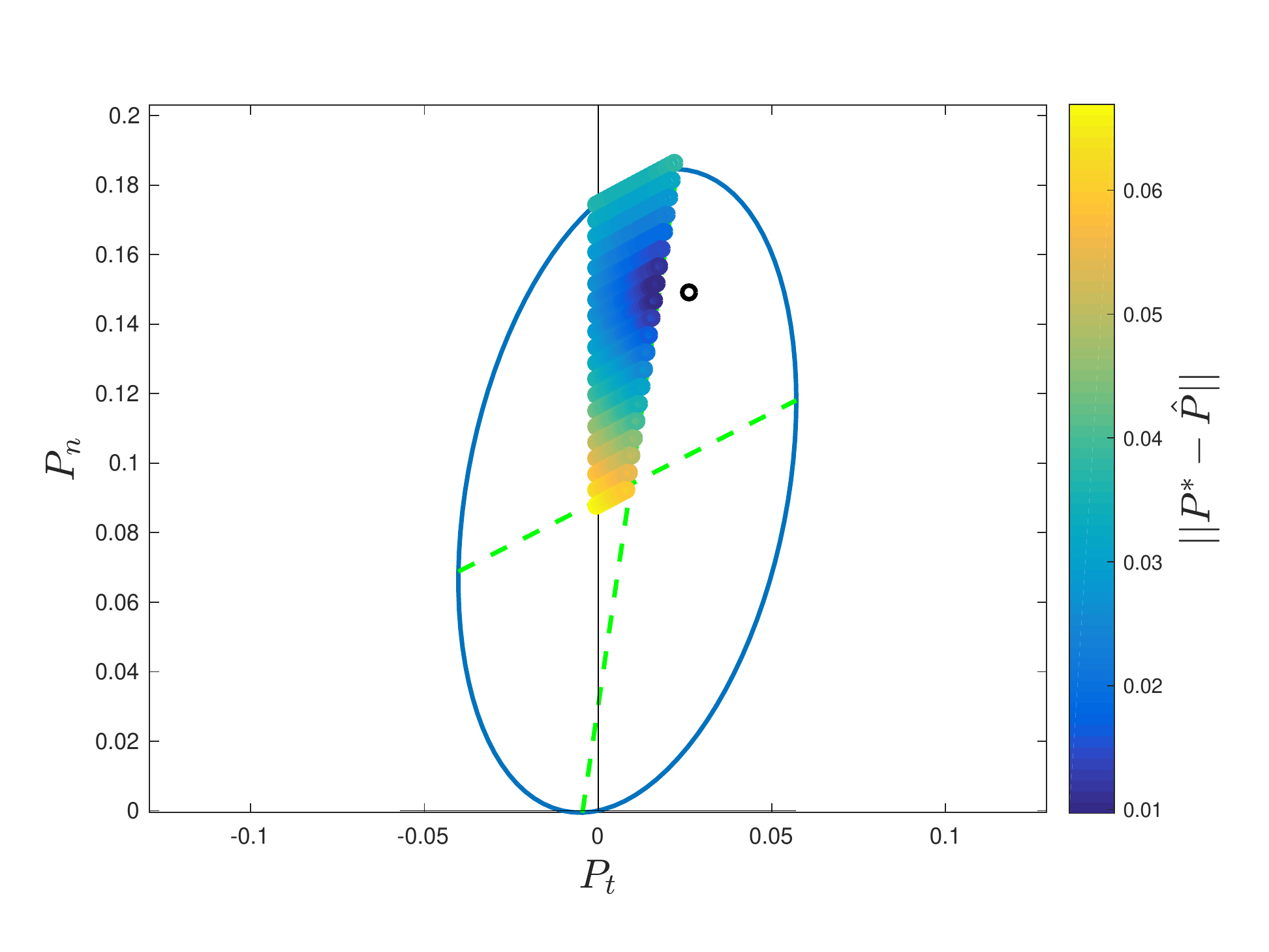}}
  \caption{a) An example of $\mu=0$, this impact is characterized by the point of contact being located directly under the center of mass of the object. In this situation, the Energy Ellipse's axis is aligned to the normal impulse direction. b) Example of a measured impulse outside the convex set of predictions made by the model due to ``spin-reversal".}
  \label{fig:vert_bounce}
\end{figure}

We now look at the anomalous cases for which $\mu=0$ in isolation; Fig.~\ref{fig:vert_bounce}a shows such an example. When the contact point lies directly below the center of mass of the object, the Energy ellipse's major axis aligns with the normal impulse direction and the convex set of predictions for the outcome of the contact collapses to a very narrow band, resulting in a very small predictive range. A small prediction range coupled with sensor noise the predictive range of the models. In the case shown in Fig.~ \ref{fig:vert_bounce}, the space spans to the left, but the measured impulse lies to the right; consequently, the value of the coefficient of friction tends to zero to get as close as possible to the target value. This issue highlights the difficulties in interpreting $\mu$ as the coefficient of friction within the context of these impact models.
\vspace{-0.5cm}
\vspace{-0.35cm}
The second set of anomalous cases is characterized by the inability to predict ``spin-reversal", an example of which can be seen in Fig.~\ref{fig:vert_bounce}b). In these scenarios, the measured impulse does not fall within the predictive range of the model and a residual error will always be present, thereby explaining the region in Fig.~\ref{fig:best_individual} where the parameters take on values with non-zero errors. Tab.~\ref{tab:conv_check} quantifies the fraction of data that lies within the convex set of each model. The relative ratios are quite low which indicates that the models suffer from potentially restrictive prediction spaces. Note that models with similar regions in the energy ellipse exhibit similar ratios, exemplified by the AP Poisson and the Drumwright-Shell models.

\vspace{-0.5cm}
\begin{table}[]
    \centering
    \setlength\tabcolsep{0.2cm}
    \caption{Fraction of drops that lie within the predictive range of each model}
    \begin{tabular}{| c | c | c | c | c | c | c |} \hline 
   &  
        DrumShell       &  AP Poisson & AP Newton & Mirtich & Wang-Mason & Whittaker \\ \hline 
      \textbf{Fraction} & 0.355 & 0.357 & 0.458 & 0.413 & 0.455 & 0.456 \\ \hline
    \end{tabular}
    \label{tab:conv_check}
\end{table}

\vspace{-1.35cm}
\section{Conclusion and Summary}
\vspace{-0.3cm}
Rigid contact models are popular due to their simplicity and computational efficiency yet they exhibit clear limitations in predictive power and accuracy. In this paper, we empirically demonstrated the inability to predict back-spin and challenges with both system identification and interpretation of model parameters.

The limitations of these analytical models can partially be attributed to the parameterization of a subset of the admissible impulse space. The independence of the parameters from the inputs to the models and the limited regions for prediction (Tab.~\ref{tab:conv_check}) contribute to the errors in predictions. Two approaches can remedy these shortcomings, we may either learn the mapping directly from the impulses and do away with the parameterization all together (data-driven approach) or use the model predictions as a prior and correct the impulse predictions using a data-driven model (data-augmented approach). In \citet{fazeli2017learning} we explore these approaches and compare performances across analytical and learned models.

The rigid models studied assume a single point contact and instantaneous impact. This assumption is only an approximation and the impact event occurs over a finite time and involves a small amount of local deformation leading to a distributed impact impulse over a patch. In \citet{fazeli2017learning} we develop a data-driven contact model that accounts for this deformation using an instantaneous wrench and demonstrate that the resulting model can significantly outperform the rigid contact models. 

This study demonstrates the importance of data in model validation and highlights the importance of experimentation in parameter selection, and its potential role in developing data-driven contact models \cite{fazeli2017learning,Yu2016}. We hope that this study serves as motivation to experimentally validate other models frequently leveraged in robotics and leads to their better understanding.


\vspace{-0.7cm}
\small{\bibliography{nf-isrr-2017.bib}}

\end{document}